\documentclass[conference]{IEEEtran}
\IEEEoverridecommandlockouts
\makeatletter
\newcommand{\linebreakand}{%
  \end{@IEEEauthorhalign}
  \hfill\mbox{}\par
  \mbox{}\hfill\begin{@IEEEauthorhalign}
}
\makeatother
\usepackage{cite}
\usepackage{amsmath,amssymb,amsfonts}
\usepackage{algorithmic}
\usepackage{graphicx}
\usepackage{subcaption}
\usepackage{textcomp}
\usepackage{xcolor}
\usepackage{booktabs}
\def\BibTeX{{\rm B\kern-.05em{\sc i\kern-.025em b}\kern-.08em
    T\kern-.1667em\lower.7ex\hbox{E}\kern-.125emX}}
\begin{document}

\title{Runway vs. Taxiway: Challenges in Automated Line Identification and Notation Approaches}

\author{\IEEEauthorblockN{Parth Ganeriwala}
\IEEEauthorblockA{\textit{Department of Computer Science} \\
\textit{Florida Institute of Technology}\\
Melbourne, USA \\
pganeriwala2022@my.fit.edu}
\and
\IEEEauthorblockN{Amy Alvarez}
\IEEEauthorblockA{\textit{Department of Computer Engineering} \\
\textit{Florida Institute of Technology}\\
Melbourne, USA \\
alvareza2023@my.fit.edu}
\and
\IEEEauthorblockN{Abdullah AlQahtani}
\IEEEauthorblockA{\textit{Department of Computer Engineering} \\
\textit{Florida Institute of Technology}\\
Melbourne, USA \\
aalqahtani2022@my.fit.edu}
\and
\IEEEauthorblockN{Siddhartha Bhattacharyya}
\IEEEauthorblockA{\textit{Department of Computer Science} \\
\textit{Florida Institute of Technology}\\
Melbourne, USA \\
sbhattacharyya@fit.edu}
\and
\IEEEauthorblockN{Mohammed Abdul Hafeez Khan}
\IEEEauthorblockA{\textit{Department of Computer Science} \\
\textit{Florida Institute of Technology}\\
Melbourne, USA \\
mkhan@my.fit.edu}
\and
\IEEEauthorblockN{Natasha Neogi}
\IEEEauthorblockA{\textit{Langley Research Center} \\
\textit{NASA}\\
Hampton VA, USA \\
natasha.a.neogi@nasa.gov }
}

\maketitle

\begin{abstract}
The increasing complexity of autonomous systems has amplified the need for accurate and reliable labeling of runway and taxiway markings to ensure operational safety. Precise detection and labeling of these markings are critical for tasks such as navigation, landing assistance, and ground control automation. Existing labeling algorithms, like the Automated Line Identification and Notation Algorithm (ALINA), have demonstrated success in identifying taxiway markings but encounter significant challenges when applied to runway markings. This limitation arises due to notable differences in line characteristics, environmental context, and interference from elements such as shadows, tire marks, and varying surface conditions. To address these challenges, we modified ALINA by adjusting color thresholds and refining region of interest (ROI) selection to better suit runway-specific contexts. While these modifications yielded limited improvements, the algorithm still struggled with consistent runway identification, often mislabeling elements such as the horizon or non-relevant background features. This highlighted the need for a more robust solution capable of adapting to diverse visual interferences. In this paper, we propose integrating a classification step using a Convolutional Neural Network (CNN) named AssistNet. By incorporating this classification step, the detection pipeline becomes more resilient to environmental variations and misclassifications. This work not only identifies the challenges but also outlines solutions, paving the way for improved automated labeling techniques essential for autonomous aviation systems.
\end{abstract}


\section{Introduction}
The development of fully autonomous vehicles has made significant advancements in computer vision and robotics. Camera-based lane and line detection systems play a crucial role in enabling vehicles to perceive the surroundings and navigate safely \cite{Cultrera_2020_CVPR_Workshops, ganeriwalacross2023}. Within the aviation industry, accurate runway detection is paramount for safe landings and takeoffs, particularly under challenging conditions like poor weather and low visibility. Accidents occurring during the taxiing phase, often attributed to factors like crew distractions and communication breakdowns with Air Traffic Control (ATC), highlight the need for automated perception systems that can augment pilot awareness and decision making on the ground. While some research has explored light-based guidance systems and other sensor modalities (e.g., LIDAR), computer vision-based approaches for taxiway line detection remain limited. Early efforts focused on grayscale images and nose-wheel mounted cameras \cite{benders2022automated}, while more recent work has incorporated color images \cite{eaton2015image} and explored camera placement on the aircraft’s vertical stabilizer \cite{Batra2020}. However, these approaches have been primarily tested in simulated environments or under ideal conditions, and their performance in real-world scenarios remains a challenge \cite{gaikwad2023developing}.  

This research builds on the work of Ganeriwala et al. \cite{ganeriwala2023assisttaxi} and Khan et al. \cite{khan2024alina} by enhancing the automated labeling process, specifically through the ALINA annotation framework \cite{khan2024alina}. ALINA was developed to detect and label taxiway line markings in video frames by establishing a consistent trapezoidal region of interest (ROI) in the initial frame, which is maintained throughout the video sequence. After applying geometric adjustments and color space transformations, ALINA generates a binary pixel map. The framework utilizes the CIRCular threshoLd pixEl Discovery And Traversal (CIRCLEDAT) algorithm to identify pixels representing taxiway markings, producing annotated frames and coordinate data files. ALINA was validated on 60,249 frames from the AssistTaxi dataset \cite{ganeriwala2023assisttaxi}, focusing on three videos with different camera perspectives, establishing a rigorous foundation for ALINA's effectiveness and its potential scalability to larger datasets.

However, while ALINA has demonstrated success with taxiway line detection, it has not yet been adapted for runway markings. Runway and taxiway line characteristics differ significantly in terms of environmental context, marking types, and structural layouts. This paper addresses the challenges in extending automated line detection algorithms to runway markings, highlighting the need for a classification step to distinguish between runway and taxiway surfaces before applying labeling algorithms. Initial modifications to ALINA, including adjustments to color thresholds and manual ROI selection, have shown limited improvements, with the algorithm often mislabeling background elements such as the horizon. Future adaptations, such as dynamic ROI adjustment and integration of contextual cues like horizon and runway geometry, are proposed to improve runway marking detection accuracy and robustness.

The rest of this paper is structured as follows: Section \ref{related} reviews related work, covering advancements in road and taxiway line detection, runway detection, and methods for horizon detection and contextual understanding. We also outline the challenges in runway marking detection, highlighting key limitations in current approaches and the need for improved classification methods. Section \ref{runway} discusses the initial experiments performed to adapt the ALINA framework for runway marking detection while section \ref{proposed} introduces our proposed classification approach, detailing the dataset, model architecture, and methodology used. Section \ref{results} presents the results and discussion, with a case study on a subset of the AssistTaxi dataset. Finally, Section \ref{conclusion} concludes the paper, summarizing the key findings and discussing future directions for adapting automated line identification in aviation contexts.

\section{Related Works}
\label{related}
\subsection{Road and Taxiway Line Detection}
Existing research on road and taxiway line detection has primarily utilized geometric and model-based approaches, which leverage lane marking geometry through techniques like the Hough transform or parallelogram-based ROIs for line detection and tracking \cite{andrei2022, Honda_2024_WACV}. Feature-based methods, relying on characteristics like color, texture, and gradients, have been enhanced through machine learning techniques such as Support Vector Machines (SVM) and Principal Component Analysis (PCA) to improve detection accuracy \cite{Chen2015,ding2020}. Recent work has also explored deep learning approaches using convolutional neural networks (CNNs) and capsule networks, showing promise in high-resolution aerial imagery \cite{guan2022road}.

However, adapting these approaches for aviation applications presents unique challenges due to differences in line types and environmental variability. Meymandi-Nejad et al. \cite{meymandi2020aircraft} demonstrated that algorithms originally developed for automotive applications, such as particle filters, Hough transforms, and LaneNet \cite{Neven2018towards}, are less effective in aeronautic environments, which include visual noise, variable lighting, and limited ROI defined by aircraft camera positioning. Delezenne et al. \cite{delezenne2024autonomous} extended this research by implementing a sliding window mechanism combined with airport map data, facilitating centerline identification and decision-making for autonomous UAV taxiing. This approach highlights the role of contextual data and specialized algorithms in achieving reliable taxiway navigation. Building on these insights, recent frameworks, including ALINA \cite{khan2024alina}, focus on the unique demands of aviation environments, leveraging contextual information and algorithmic adaptations to improve taxiway marking detection. The need for tailored solutions in aeronautics, including advanced vision-based navigation systems that can handle real-world variabilities, remains essential for future advancements in both manned and unmanned aviation ground operations.

\begin{figure*}[htbp]
    \centering
    \begin{subfigure}{0.4\textwidth}
        \includegraphics[width=0.95\linewidth]{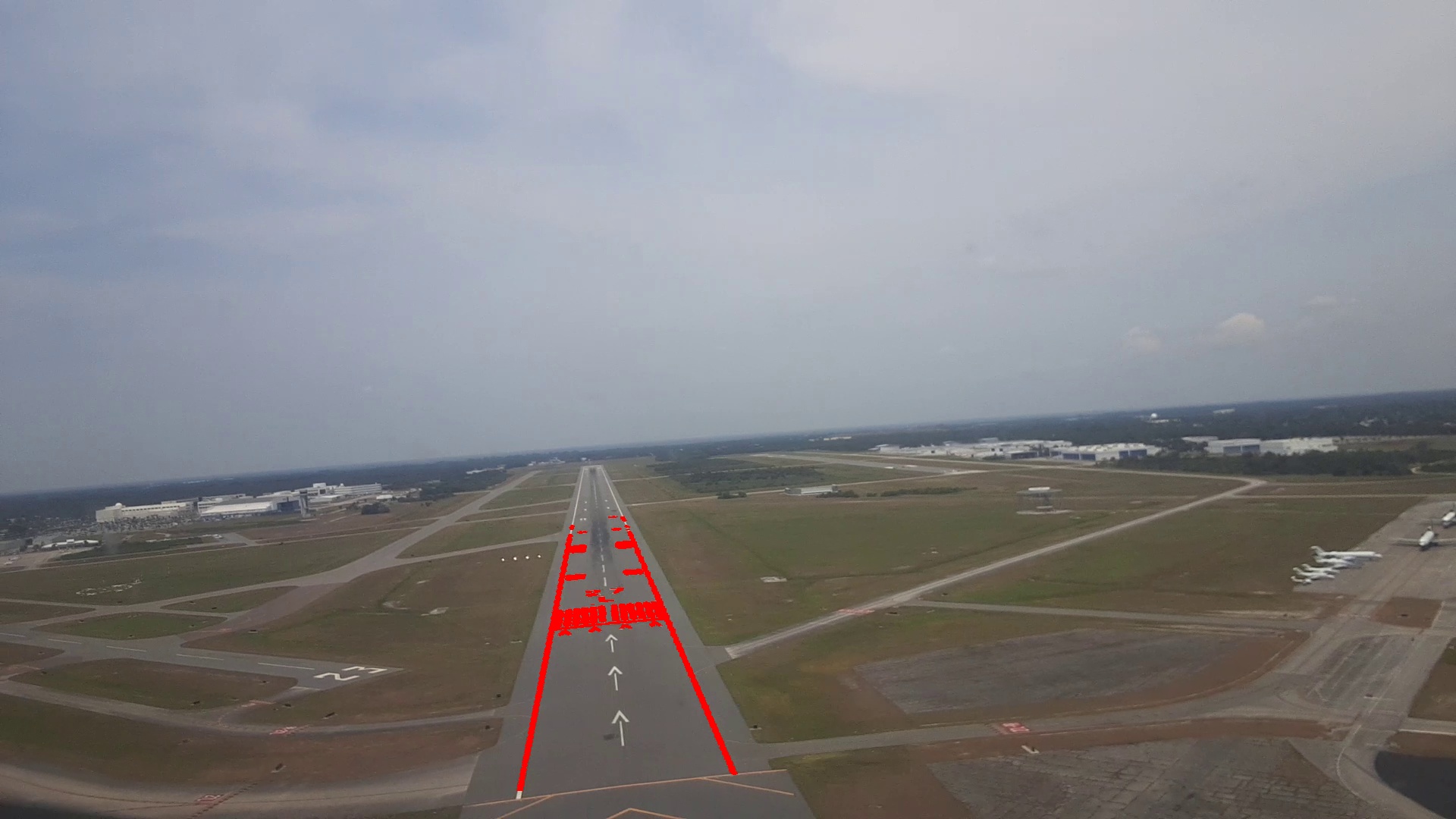}
        \caption{Accurate detection of edge lines but missing the center lines}
        \label{fig:image1}
    \end{subfigure}
    \begin{subfigure}{0.4\textwidth}
        \includegraphics[width=0.95\linewidth]{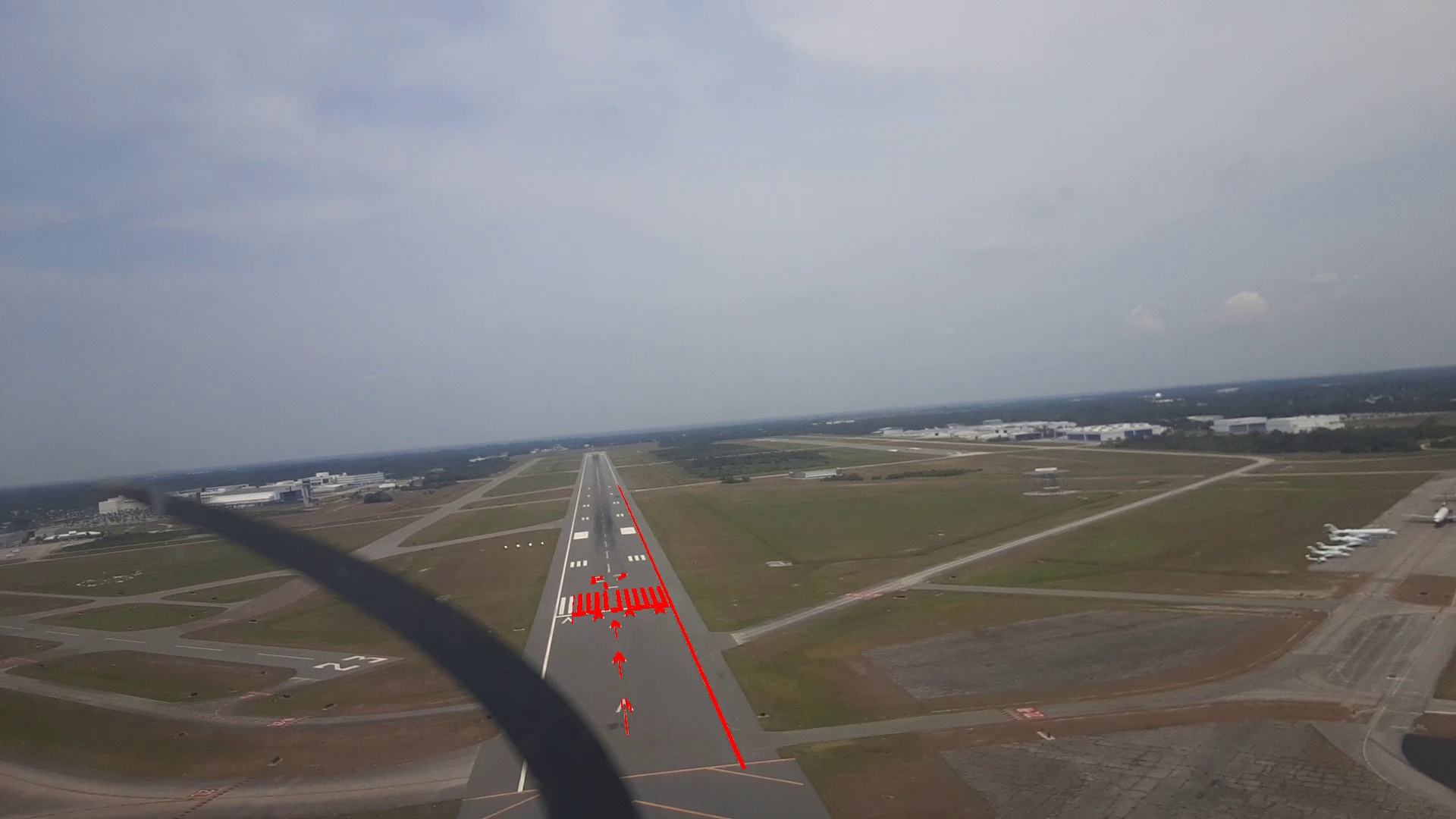}
        \caption{Partial detection achieved but with some missed segments due to propeller interference}
        \label{fig:image2}
    \end{subfigure}
    
    \vspace{0.3cm}
    \centering
    \begin{subfigure}{0.4\textwidth}
        \includegraphics[width=0.95\linewidth]{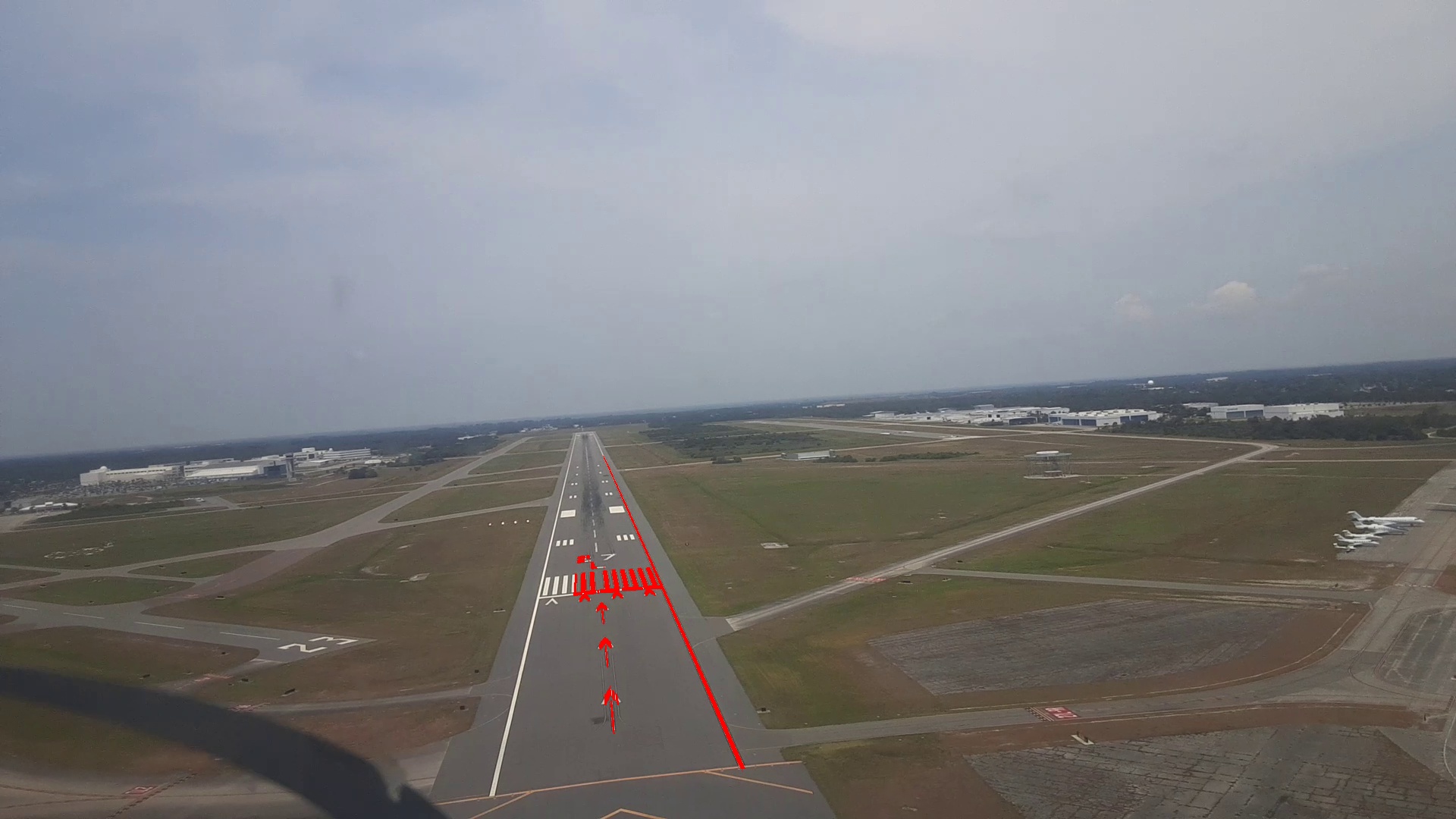}
        \caption{Partial detection achieved but with some missed segments.}
        \label{fig:image3}
    \end{subfigure}
    \begin{subfigure}{0.4\textwidth}
        \includegraphics[width=0.95\linewidth]{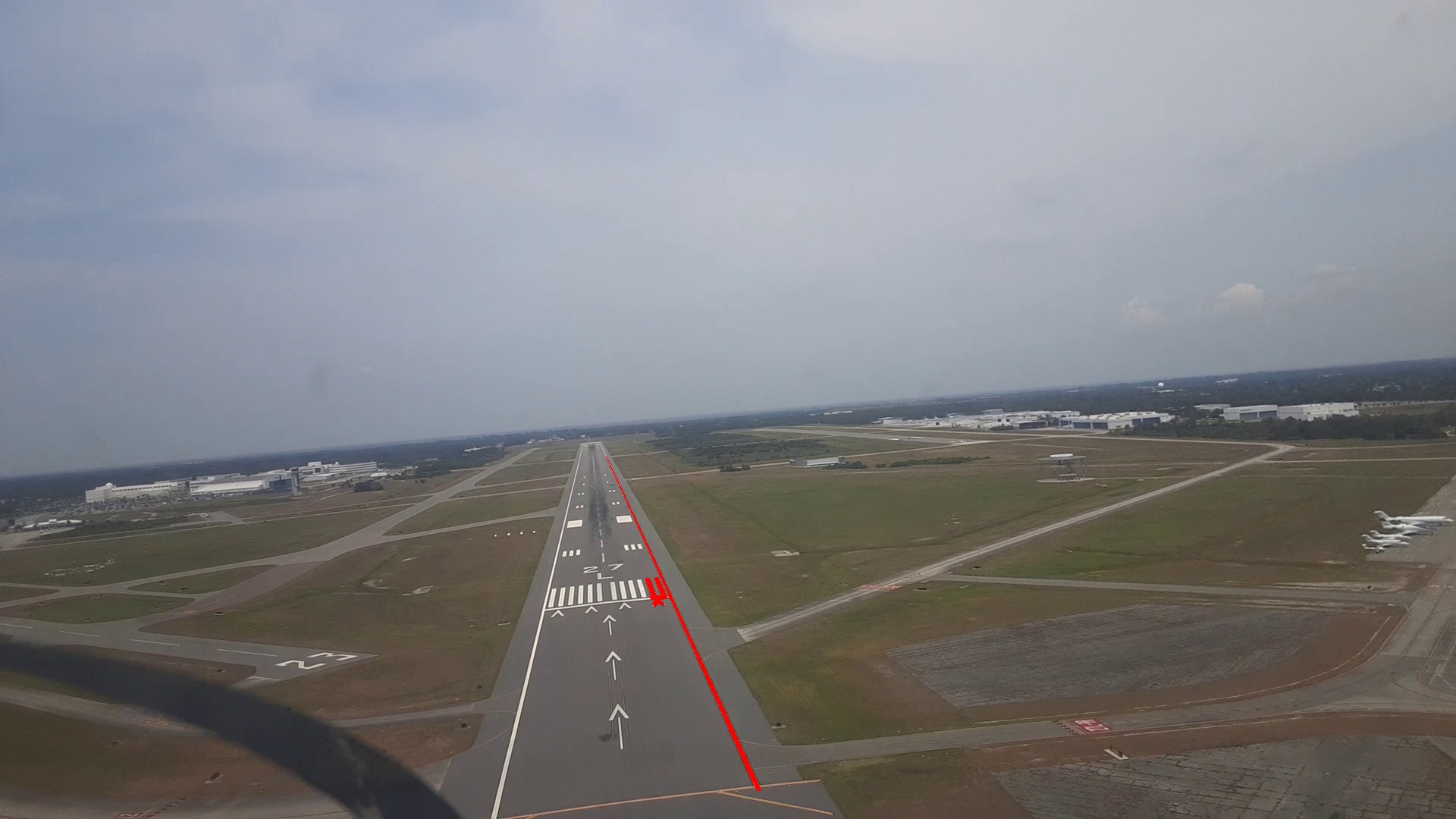}
        \caption{ALINA labels only the right line marking, with the left line missing}
        \label{fig:image4}
    \end{subfigure}
    \caption{Results of ALINA’s Line Labeling on Runway Instances with Adjusted Color Thresholding.}
    \label{fig:four_images}
\end{figure*}
\subsection{Runway Detection}
Runway detection research is less developed compared to road and taxiway line detection, with most approaches focusing on onboard camera systems \cite{ducoffe2023lard, chen2024image, li2024federated}. Traditionally, cameras mounted on the aircraft’s vertical stabilizer or nose wheel have been employed to detect runway centerlines and markings; however, these methods often rely on simplifying assumptions about line geometry and struggle under varying environmental conditions \cite{Batra2020}. Fusion with additional sensor modalities, such as LIDAR and radar, has also been investigated to improve detection reliability but generally requires added hardware and complex processing pipelines.
Amit et al. \cite{amit2021robust} proposed a two-stage R-CNN-based framework tailored to runway detection in complex aerial scenes. Their approach leverages region proposal networks (RPN) and multiscale feature extraction via DarkNet-53 to address the challenges of background clutter and variable imaging angles, achieving high detection accuracy even in diverse conditions. To address lightweight needs for embedded platforms, Dai et al. \cite{dai2024yomo} introduced YOMO-Runwaynet, which integrates YOLO and MobileNetV3 for efficient and real-time runway detection, specifically for fixed-wing aircraft. This model employs MobileNetV3 as the backbone and optimizes latency through path aggregation networks and spatial pyramid pooling, achieving high detection accuracy (89.5\%) and rapid inference speeds on embedded devices. YOMO-Runwaynet’s high speed and fault tolerance demonstrate the potential of lightweight architectures for real-time runway detection in constrained environments.

The approaches illustrate how deep learning innovations have broadened the runway detection landscape. Nevertheless, challenges remain, such as distinguishing runways from similar-looking elements like taxiways and aprons and managing extreme lighting or shadow variations.

\subsection{Horizon Detection and Contextual Understanding}

Horizon detection and contextual understanding are essential for distinguishing runway markings from visual elements like the sky, clouds, or distant landscape features. Traditional methods for horizon detection have used intensity-based approaches that capitalize on brightness contrasts to identify the horizon line \cite{boroujeni2012robust}. However, recent advancements have incorporated machine learning techniques, such as Support Vector Machines (SVM) and neural networks, to achieve more robust horizon detection under varied lighting and environmental conditions \cite{fefilatyev2006horizon,yazdanpanah2015real}.

Further progress has been made in autonomous navigation for Micro Air Vehicles (MAVs) through vision-based horizon detection. This approach draws inspiration from natural flight, observing how birds maintain stability through horizon cues. A study \cite{ettinger2002towards} on MAVs proposed a robust horizon detection algorithm capable of real-time processing at 30 Hz, achieving over 99.9\% accuracy across diverse terrains, including roads, buildings, and natural landscapes. This work highlights the foundational role of horizon detection in enabling autonomous flight stability for small, highly maneuverable aircraft, showcasing its utility in navigating complex, cluttered environments by ensuring reliable horizon line identification under variable conditions.

These advances indicate that horizon detection algorithms can provide essential contextual understanding for runway detection systems. By integrating reliable horizon lines with contextual environmental data, such as terrain features or elevation maps, computer vision-based runway detection can improve its resilience against contradicting elements like the sky or horizon, particularly in complex or low-visibility conditions.
\begin{figure*}[h!tbp]
    \centering
    \includegraphics[width=.8\linewidth]{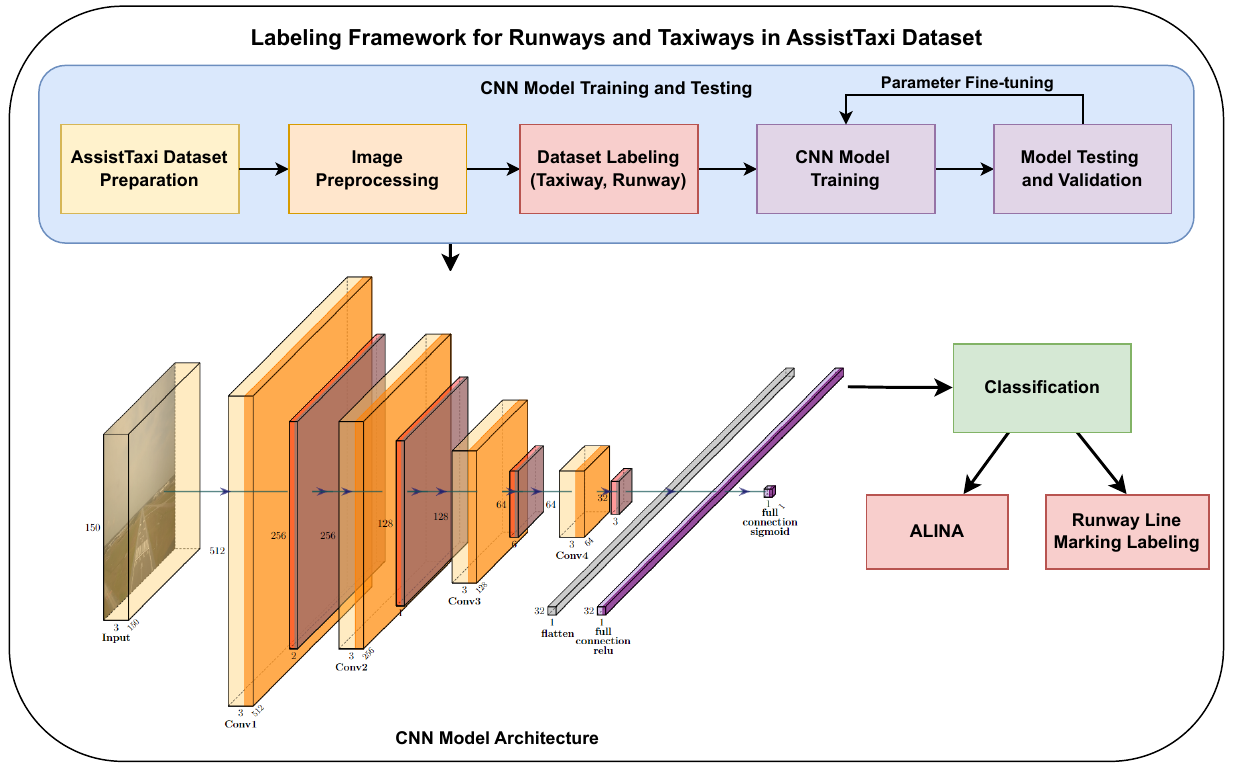}
    \caption{Proposed Labeling Framework with Initial Classification Step}\label{fig:framework}
\end{figure*}

\section{Adapting Line Identification and Notation for Runways}
\label{runway}

To address the unique challenges of runway detection, we developed refinements to the ALINA algorithm \cite{khan2024alina}, aiming to extend its utility from taxiways, where it performs effectively, to the more complex environment of runways. Current literature highlights several key obstacles in adapting taxiway labeling algorithms to runways, including the need for precise marking detection and dynamic region focus. Our approach leverages ALINA's strengths in color-based feature extraction and statistical analysis, while proposing enhancements tailored to runway environments.

One significant adaptation we explored was modifying ALINA’s color thresholding mechanism. While taxiway markings are typically yellow, runway markings are predominantly white, requiring us to adjust ALINA’s sensitivity to these color differences. By analyzing the color distribution of runway markings across diverse datasets, we fine-tuned the threshold parameters to capture white markings more accurately. Despite this adaptation, our tests indicated that this change alone was insufficient for consistently reliable runway detection, as shown in Figure \ref{fig:four_images}.

In addition to color adjustments, we recognize the importance of contextual filtering—particularly the incorporation of horizon detection and runway geometry. By leveraging this contextual information, future extensions of the algorithm could more effectively focus on the specific spatial features unique to runways, reducing the potential for false positives. Furthermore, handling varying environmental conditions such as fluctuating lighting, weather changes, and potential obstructions is essential for improving runway visibility across diverse scenarios. An additional component essential for accurate runway detection would be a dynamic ROI selection, which would allow the algorithm to focus on the most relevant areas of an image based on the aircraft’s position and orientation. Unfortunately, due to the absence of position data in our images, this component could not be implemented. However, we propose that a dynamic ROI, aligned with aircraft positioning, would enable the algorithm to adapt its focus as the aircraft approaches the runway, potentially yielding significant improvements in detection accuracy.

As ALINA performs well in detecting taxiway markings, we propose introducing an initial classification step to distinguish between taxiways and runways (Figure \ref{fig:framework}). This classification would guide the algorithm, allowing it to apply ALINA’s established labeling process to taxiways while directing a future, dedicated labeling algorithm to handle runway markings. The following section introduces our proposed classification approach, detailing the dataset, architecture, and methodology that will enable accurate initial classification of runways and taxiways.
\section{Proposed Classification Approach}
\label{proposed}
\begin{figure*}[h!tbp]
    \centering
    \begin{subfigure}{0.4\textwidth}
        \includegraphics[width=0.95\linewidth]{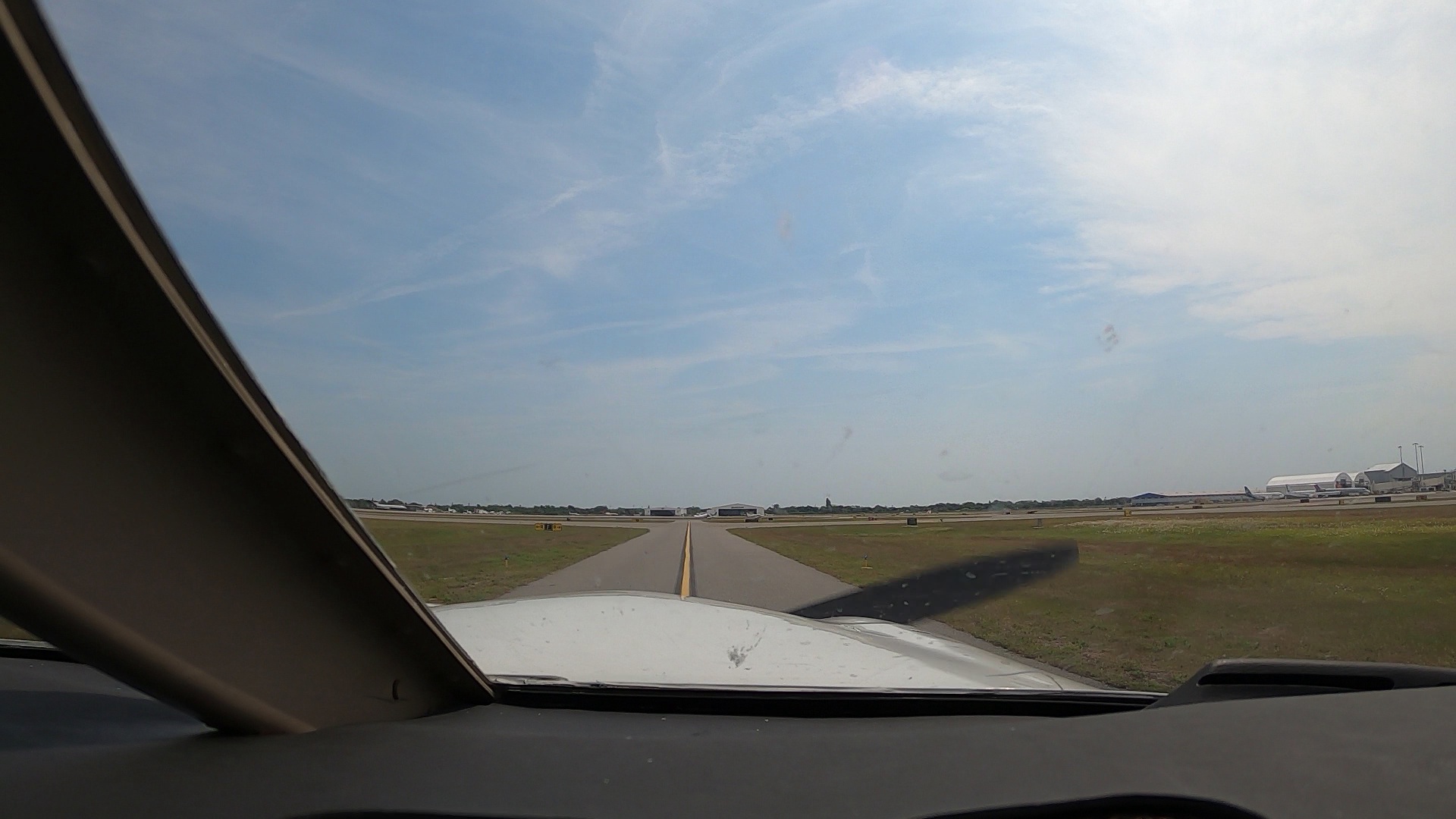}
        \caption{Taxiway Instance}\label{fig:taxiway}
        \label{fig:taxiway}
    \end{subfigure}
    \begin{subfigure}{0.4\textwidth}
        \includegraphics[width=0.95\linewidth]{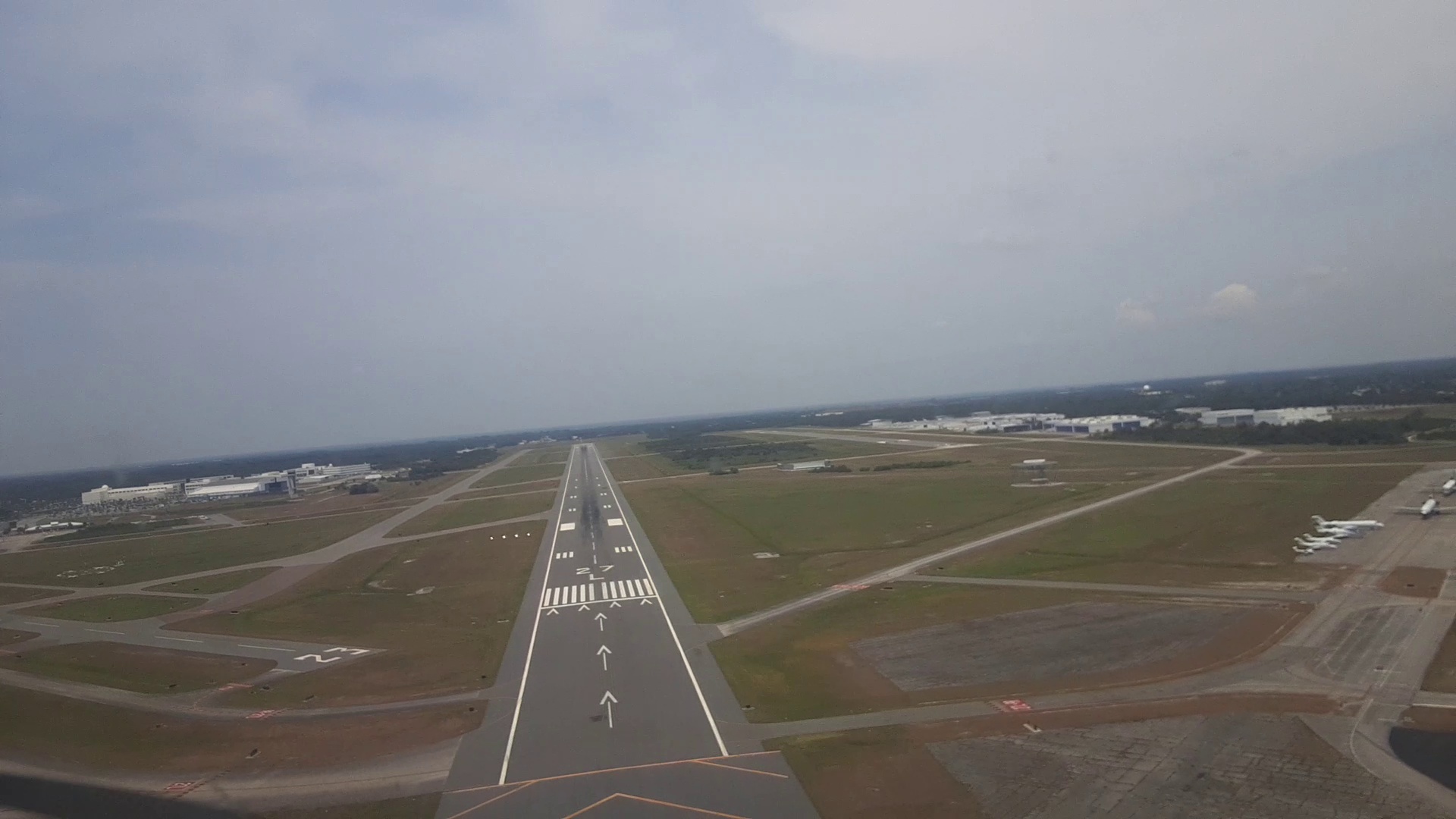}
        \caption{Runway Instance}\label{fig:runway}
        \label{fig:runway}
    \end{subfigure}    
    \caption{Instances from the AssistTaxi Dataset \cite{ganeriwala2023assisttaxi}}
    \label{fig:assistTaxi}
\end{figure*}
\subsection{Dataset}

We used a subset of the AssistTaxi dataset, a collection specifically designed for the detection of aviation surface markings, containing images from both runway and taxiway environments \cite{ganeriwala2023assisttaxi}. The AssistTaxi dataset includes data from various airport locations and captures diverse real-world conditions to enhance model robustness. For this study, we manually labeled a representative subset of images into two categories: taxiway and runway (Figure \ref{fig:taxiway} and \ref{fig:runway}). Taxiway images focus on features such as yellow surface markings, hold short signs, and directional arrows, while runway images emphasize white surface markings, threshold markings, and center lines. This labeled subset was divided into training, validation, and testing sets in a 70:20:10 ratio to ensure a balanced distribution for model development. We applied pixel normalization and basic image augmentation techniques, such as random rotations and brightness adjustments, to increase the dataset’s diversity and the model’s resilience to changes in lighting and perspective.

\subsection{Architecture}

For this classification task, we implement a Convolutional Neural Network (CNN) named \textit{AssistNet}, which is designed to distinguish between runway and taxiway images. The architecture consists of a sequence of convolutional layers, each followed by max-pooling layers, enabling the model to learn hierarchical features from the input images. We have chosen this architecture due to the inherent similarities between taxiways and runways, both of which share a road-like appearance. The key differentiating factor is the color of the line markings—yellow for taxiways and white for runways. To effectively capture these differences, the initial convolutional layers use Conv2D to apply filters that extract spatial features, allowing the model to focus on the relevant patterns associated with each class. As the layers progress, the number of filters decreases, concentrating on increasingly abstract features of the images.

After each convolutional layer, we utilize MaxPooling2D to downsample the feature maps. This step helps retain the most significant features while reducing the computational load and minimizing the risk of overfitting. Furthermore, our preprocessing steps include cropping the images to remove the horizon and isolate the road segments, enhancing the model's ability to differentiate between the two classes based solely on line color. Following the convolutional and pooling layers, a Flatten layer converts the 2D feature maps into a 1D array, making it suitable for input into the fully connected layers. The model then includes fully connected (Dense) layers, allowing the network to learn complex relationships within the data. The output layer utilizes a sigmoid activation function to classify the images as either runway or taxiway, making it appropriate for binary classification tasks. We use the binary cross-entropy loss function, which is standard for binary classification problems, along with the ADAM optimizer, known for its efficiency in converging quickly. The implementation of \textit{AssistNet}, our CNN architecture, is also available on our GitHub repository \footnote{https://github.com/AmyAlvarez/AssistNet.git}.

\subsection{Proposed Methodology }

The classification process begins by passing input images through AssistNet, our CNN. The feature extractor within the network identifies key surface patterns and contextual elements, such as the color and geometry of the lines, which are crucial for distinguishing between runway and taxiway markings. Given the similarities between the two, it is essential to focus on these specific features to achieve accurate classification. To enhance the model's performance, we incorporate an contextually reduced ROI algorithm. This technique helps narrow the focus to surface-level features by eliminating irrelevant visual information, such as the horizon and land along the runway, which can often confuse the classification process. By concentrating on the line markings themselves, we aim to improve the model's ability to differentiate between the two classes based solely on color cues—yellow lines for taxiways and white lines for runways. 

In our pre-processing steps, we first resize the input images to a consistent resolution of 400x225 pixels. This resizing ensures uniformity across all input data. Additionally, we apply further enhancements, including normalization of pixel values to standardize the brightness and contrast across images. This is complemented by image augmentation techniques, such as random rotations and brightness adjustments, which help to simulate various environmental conditions and enhance the robustness of the model. By utilizing these methodologies, we aim to create a comprehensive approach to classifying runway and taxiway images accurately. The combination of targeted feature extraction, reduced ROI selection, and comprehensive preprocessing forms the foundation of our proposed classification system.


\section{Results and Discussion}
\label{results}

\begin{figure}[h!tbp]
    \centering
    \includegraphics[width=\linewidth]{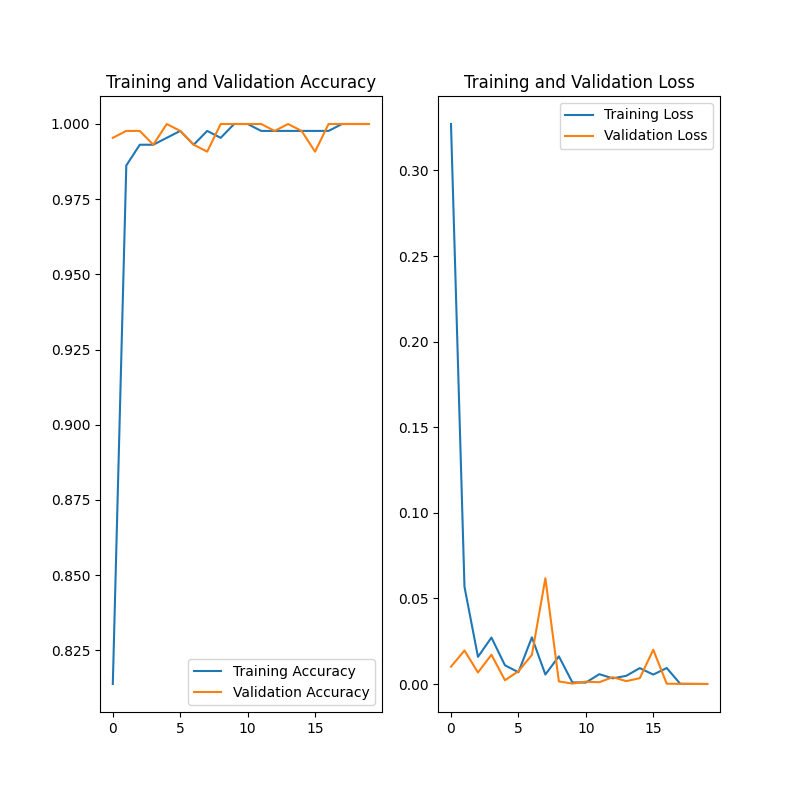}
    \caption{Training and Validation Results over 20 epochs}
    \label{fig:train}
\end{figure}

This section presents the results obtained (see Figure \ref{fig:train}) from the classification experiments using AssistNet on the AssistTaxi dataset \cite{ganeriwala2023assisttaxi}. Our primary aim was to evaluate the model's performance in differentiating between runway and taxiway images while addressing various challenges, including visual interferences such as surface markings, threshold markings, and center lines. The experiments were conducted on a system equipped with an NVIDIA GTX 3060TI GPU, 16 GB of RAM, a Ryzen 9 AMD CPU, TensorFlow 2.0, and Keras. 

\begin{figure*}[h!tbp]
    \centering
    \includegraphics[width=0.8\linewidth]{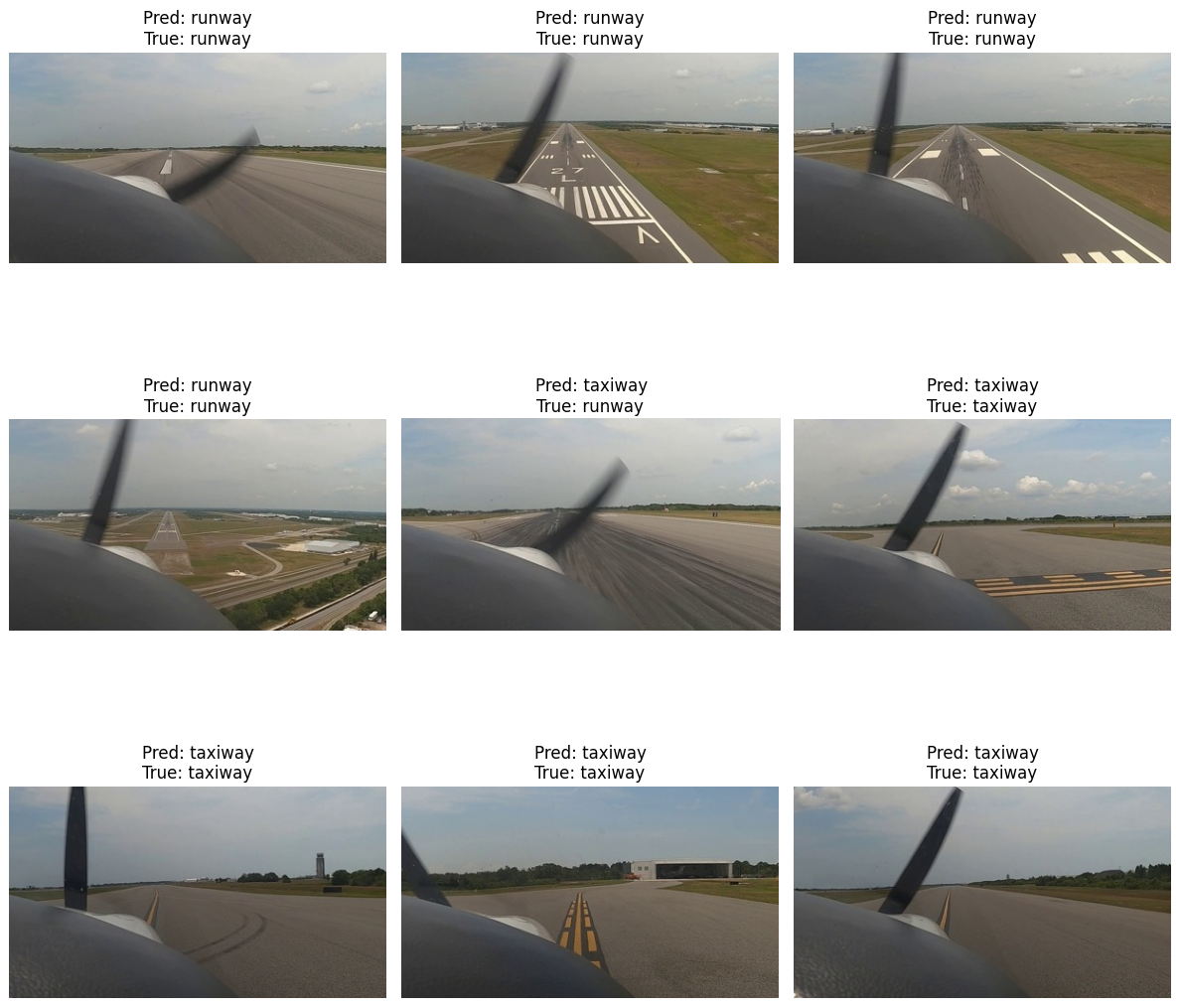}
    \caption{Classification Results}
    \label{fig:enter-label}
\end{figure*}
Initially, we used 500 images, comprising 250 runway and 250 taxiway samples, which were further divide into 125 images for training and validation for runways, along with 125 images for training and validation for taxiways. These images were captured from various camera angles and fields of view, including during takeoff and departure operations. The selected images represented challenging conditions, such as deteriorated markings due to wear, blurred areas from motion, and overlapping signs. Each image was resized to 150x150 pixels for consistency across the dataset. To improve generalization and reduce the occurrence of overfitting, data augmentation techniques were applied, including various random rotations and brightness adjustments. This augmentation ensured that the model was exposed to a variety of image transformations, improving its ability to generalize to unseen data.

Initially, the model was trained on a limited dataset, under the assumption that the relatively straightforward visual distinction between runway and taxiway markings would be sufficient. However, initial performance evaluations revealed a consistent misclassification issue due to the presence of visually similar textures and environmental interferences in the surrounding road surface, which the model struggled to differentiate. This indicated that the dataset's limited scope lacked sufficient variability to capture these nuances, impacting the model's generalization capabilities.

To address these limitations, we adopted an iterative approach focused on dataset expansion and data preprocessing refinement. First, the dataset size was significantly increased to introduce greater variability by incorporating images captured from a broader range of camera angles, lighting conditions, and scenarios involving partial obscurations of markings. This expansion provided a more comprehensive representation of operational conditions, leading to an observed improvement in classification performance, although issues persisted with highly similar textures in specific contexts.

Following this, a contextually reduced ROI extraction method was implemented to enhance the model's focus on relevant features. By isolating image sections containing only the markings, we effectively minimized peripheral interference from non-essential elements such as surrounding pavement and sky. Additionally, the image resolution was increased to 400x225 pixels, with particular emphasis on preserving high-contrast areas to maintain detail. This iterative refinement of data preprocessing enabled the model to concentrate more effectively on distinguishing essential features, resulting in improved classification accuracy. Each adjustment cycle was evaluated through validation testing, with the resulting insights guiding further refinements to the dataset and model architecture.


The model demonstrated an ability to classify images accurately, achieving an overall accuracy of 99.5\% on the validation dataset. The confusion matrix revealed that the model effectively identified runway markings with a high precision rate, although some confusion occurred with taxiway markings under certain conditions, such as when markings were partially obscured or faded. We also evaluated the model's performance under challenging conditions. For instance, images captured during takeoff and departure, which included motion blur and overlapping features, posed difficulties; however, AssistNet maintained satisfactory accuracy levels. The use of the reduced ROI algorithm significantly contributed to the model's robustness by focusing on surface-level features and minimizing the impact of irrelevant visual information. The data augmentation strategies applied during preprocessing played a crucial role in enhancing the model's generalization capability. By simulating various environmental conditions through techniques like random rotations and brightness adjustments, we observed improved performance on unseen data. The experiments also underscored the importance of a balanced dataset, as it facilitated the model's learning process by providing a diverse range of examples for both classes. 


In future work, we aim to further refine AssistNet's architecture and explore more advanced augmentation techniques, such as varying weather conditions and different lighting scenarios, to improve its performance in real-world applications. 

\section{Limitations of Current Runway Marking Detection Systems}
The adaptation of existing line detection algorithms to runway environments presents several challenges such as:
\begin{itemize}
    \item Adaptive ROI: The need for a dynamic ROI \cite{yoo2021graph} that adjusts to the aircraft’s changing perspective and the varying visibility of the runway during approach. 
    \item Contextual Filtering: The importance of incorporating contextual information like horizon detection and runway geometry to reduce false positives and improve detection accuracy. 
    \item Robustness to Environmental Variations: The need to handle changing lighting conditions, weather patterns, and potential obstructions that may impact runway visibility \cite{tsapparellas2023vision}. 
\end{itemize}
Future research should focus on addressing these challenges and exploring the potential of multi-modal sensor fusion to enhance runway detection reliability under diverse and demanding conditions.

\section{Conclusion}
\label{conclusion}
In this study, we addressed the challenge of accurately labeling runway and taxiway markings in the context of autonomous systems. Our findings highlight the limitations of existing automated line detection algorithms, particularly ALINA, when applied to runway markings, which exhibits distinct characteristics compared to taxiway markings. Despite implementing initial modifications aimed at enhancing ALINA's performance—such as refining color threshold settings and optimizing ROI selection—we found that the algorithm's efficacy was still limited. This was particularly due to the confusion caused by environmental elements such as the horizon. We therefore proposed implementing an initial classification step to differentiate between runway and taxiway surfaces before deploying labeling algorithms. This approach not only improves the accuracy of marking identification but also reduces the potential for mislabeling. We propose several avenues for future research which include the integration of dynamic ROI adjustments and contextual cues, such as horizon detection and runway geometry. Furthermore, exploring sensor fusion techniques that incorporate data from LiDAR and radar could significantly enhance detection capabilities. In conclusion, our work lays the groundwork for advancing automated runway marking detection and labeling processes, contributing to the broader goal of enhancing the operational safety and efficiency of autonomous systems in aviation.
\bibliographystyle{plain}
\bibliography{refs}
\end{document}